\documentclass[10pt,twocolumn,letterpaper]{article}

\usepackage[pagenumbers]{cvpr} 

\definecolor{cvprblue}{rgb}{0.21,0.49,0.74}
\usepackage[pagebackref,breaklinks,colorlinks,allcolors=cvprblue]{hyperref}

\usepackage[accsupp]{axessibility} 
\usepackage{bbding}
\usepackage{bm}
\usepackage{colortbl}
\usepackage{tcolorbox}
\tcbuselibrary{skins, breakable}

\definecolor{Gray}{gray}{0.90}

\newcommand{\sref}[1]{\S\ref{#1}}
\newcommand{\sssection}[1]{\noindent\textbf{#1}}

\newcommand{\method}{Bind \& Compose}
\newcommand{\methodshort}{BiCo}

\def\paperID{****} 
\def\confName{CVPR}
\def\confYear{2026}

\title{Composing Concepts from Images and Videos via Concept-prompt Binding}

\author{Xianghao Kong$^1$~, Zeyu Zhang$^1$~, Yuwei Guo$^2$~, Zhuoran Zhao$^{1,3}$~, Songchun Zhang$^1$~, Anyi Rao$^1$\\
\small{$^1$~HKUST} \ 
\small{$^2$~CUHK} \ 
\small{$^3$~HKUST(GZ)} \\
\url{https://refkxh.github.io/BiCo_Webpage}
}

\begin{document}

\twocolumn[{
\renewcommand\twocolumn[1][]{#1}
\maketitle
\vspace{-2em}
\includegraphics[width=\linewidth]{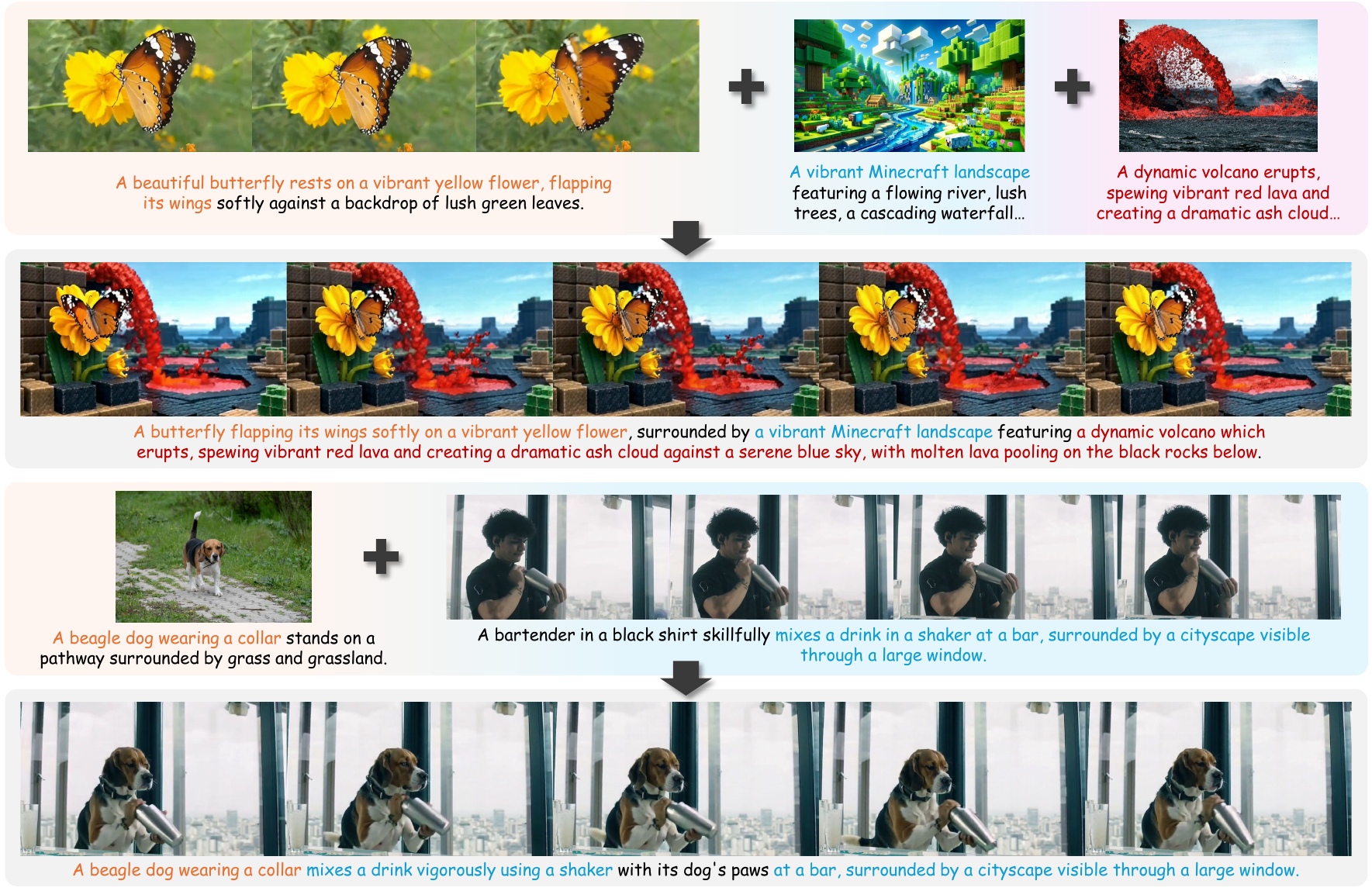}
\captionof{figure}{Illustration of \methodshort , a one-shot method that enables flexible visual concept composition by binding visual concepts with the corresponding prompt tokens and composing the target prompt with bound tokens from various sources (\sref{sec:intro}).
\vspace{1em}
}
\label{fig:teaser}
}]

\begin{abstract}
Visual concept composition, which aims to integrate different elements from images and videos into a single, coherent visual output, still falls short in accurately extracting complex concepts from visual inputs and flexibly combining concepts from both images and videos.
We introduce \method , a one-shot method that enables flexible visual concept composition by binding visual concepts with corresponding prompt tokens and composing the target prompt with bound tokens from various sources. 
It adopts a hierarchical binder structure for cross-attention conditioning in Diffusion Transformers to encode visual concepts into corresponding prompt tokens for accurate decomposition of complex visual concepts.
To improve concept-token binding accuracy, we design a Diversify-and-Absorb Mechanism that uses an extra absorbent token to eliminate the impact of concept-irrelevant details when training with diversified prompts. To enhance the compatibility between image and video concepts, we present a Temporal Disentanglement Strategy that decouples the training process of video concepts into two stages with a dual-branch binder structure for temporal modeling. Evaluations demonstrate that our method achieves superior concept consistency, prompt fidelity, and motion quality over existing approaches, opening up new possibilities for visual creativity. 
\vspace{-1.5em}
\end{abstract}    
\section{Introduction}
\label{sec:intro}



Visual concept composition aims to integrate different elements from images and videos into a single, coherent visual output. This process is a reflection of human artists' creation: combining ingredients from various inspirations to form a brand new masterpiece~\cite{ipcomposer}. Consequently, it plays a fundamental role in visual creativity and filmmaking~\cite{film_survey}. With the rapid advancement of diffusion-based visual content generation models~\cite{sd3, labs2025flux1kontextflowmatching, easyanimate, cogvideox, opensora2, hunyuanvideo, hacohen2024ltxvideorealtimevideolatent, wan, data_synth_3dmm, musetalk, kong2025profashion, ifv2v}, an increasing number of works~\cite{abdal2025zeroshotdynamicconceptpersonalization, dynamicconcept_singlevideo, tokenverse, videomage, dreamvideo, wang2025dualrealadaptivejointtraining, kwontweediemix, chen2025xverseconsistentmultisubjectcontrol, dalva2025lorashoptrainingfreemulticonceptimage, omg, conceptweaver} have been exploring the field of visual concept composition by exploiting the generative models' strong capability of concept grounding and customization. 


Despite considerable efforts devoted to this field, challenges still remain in accurately extracting complex concepts from visual inputs and flexibly combining concepts from both images and videos.
First, the capability to precisely extract specific concepts from various sources is of great significance for visual content creators. 
Nevertheless, existing mainstream methods~\cite{abdal2025zeroshotdynamicconceptpersonalization, dynamicconcept_singlevideo, videomage, dreamvideo, wang2025dualrealadaptivejointtraining, dalva2025lorashoptrainingfreemulticonceptimage, omg, conceptweaver} use either adapters like LoRA~\cite{lora} or learnable embeddings with explicit or implicit masks to realize concept selection, which fall short in decoupling complex concepts with occlusions and temporal alterations, and extracting non-object concepts such as styles.
Second, it is a common practice to integrate different visual elements from both images and videos in the visual content creation process~\cite{film_survey}. However, previous works are confined to animating designated subjects from images with motion from videos~\cite{videomage, dreamvideo, wang2025dualrealadaptivejointtraining}, without further exploration of flexibly combining various attributes (\eg, visual styles and lighting variations) from both images and videos. Although there has been recent effort on flexible concept composition~\cite{tokenverse} in the image domain, achieving universal visual concept composition for both images and videos remains an underexplored problem. 



To this end, we introduce \method ~(\methodshort), a one-shot method that enables flexible visual concept composition by binding visual concepts with the corresponding textual tokens, with satisfactory compatibility between image and video concepts (\cref{fig:teaser}). 
Our method first leverages the powerful concept grounding capability~\cite{diffpng} of text-to-video (T2V) diffusion models~\cite{wan} to bind textual tokens with their corresponding visual concepts through one-shot training, achieving implicit decomposition without mask input. 
Then, concept composition is done through selecting any desired bound tokens from various sources and composing them into the final prompt tokens, which serves as the model condition. 
This paper mainly encompasses the following three technical contributions:
\textbf{First}, to achieve reliable decomposition of complex visual concepts for flexible manipulation, we propose a hierarchical binder structure for the cross-attention mechanism~\cite{attention2017} in Diffusion Transformer (DiT)~\cite{dit} blocks to effectively encode visual concepts into corresponding textual tokens. 
When composing concepts from multiple sources, concept tokens in the target prompt are passed through different binders correspondingly to integrate visual features, enabling text-conditioned concept composition without explicit mask input.
\textbf{Second}, to improve the accuracy of concept-token binding for more precise concept decomposition, we design a Diversify-and-Absorb Mechanism (DAM) that diversifies the one-shot prompts while retaining key concepts, and introduces an extra absorbent token during training to eliminate the impact of concept-irrelevant details. \textbf{Third}, to enhance the compatibility between image and video concepts during composition, we present a Temporal Disentanglement Strategy (TDS) that decouples the training process of video concepts into two stages. In the first stage, the binders are trained with individual frames without temporal concepts, which aligns with the training setting of image concepts. In the second stage that trains the binders on videos, we adopt a dual-branch binder structure to better cater to temporal concepts while inheriting knowledge from the first stage. 

Extensive experiments demonstrate that \methodshort ~simultaneously achieves superior concept consistency, prompt fidelity, and motion quality when performing visual concept composition. It also outperforms previous baseline approaches in both concept manipulation flexibility and visual quality of the composed video. With support for a variety of innovative video creation tasks, \methodshort ~demonstrates a strong potential to serve as a promising solution for creators to experiment with their whimsies.

\section{Related Work}
\label{sec:related}

\sssection{T2V Diffusion Models.} The emergence of diffusion models~\cite{ddpm, ddim, ldm} has significantly advanced the realm of visual content generation. Recently, DiT~\cite{dit} has become the de facto standard of the denoising model's architecture, surpassing U-Net~\cite{unet} with its strong scaling capability~\cite{kaplan2020scalinglawsneurallanguage} and flexibility for integrating multi-modal conditions~\citep{sd3}. Flow Matching~\cite{flowmatching, rectifiedflow} introduces a new linear paradigm to transit between the Gaussian distribution and the target distribution, improving the theoretical properties and simplifying the conceptual framework. Based on these works, a number of T2V diffusion models~\cite{easyanimate, cogvideox, opensora2, hunyuanvideo, hacohen2024ltxvideorealtimevideolatent, wan} have emerged with text-to-video cross-attention or joint attention for conditioning. Despite these methods achieving satisfactory quality and consistency of generation, they are designed for general T2V generation, with limited support for personalization and concept composition.

\begin{figure*}[t]
    \centering
    \includegraphics[width=0.93\linewidth]{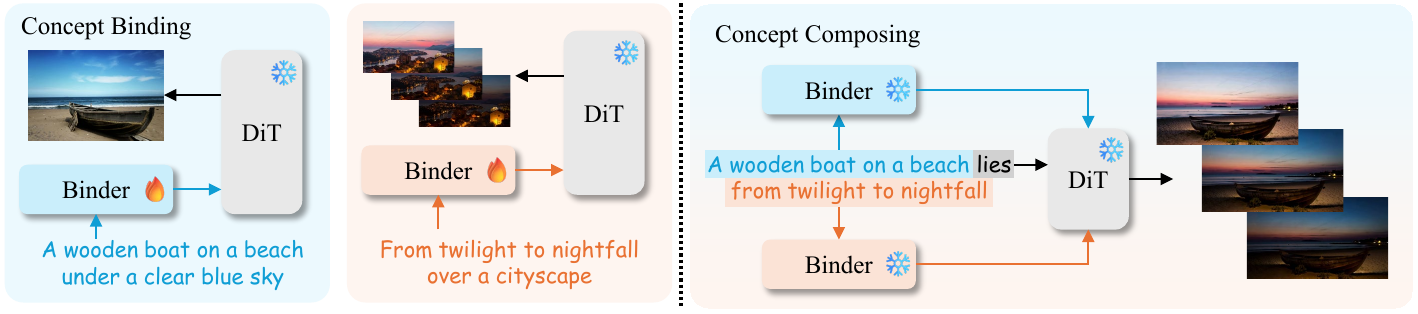}
    \vspace{-0.5em}
    \caption{\textbf{Overview of \methodshort} (\sref{sec:method_overview}). \methodshort ~first adopts a binder structure to learn visual concepts into corresponding prompt tokens, and then composes different concepts by passing corresponding prompt tokens through different adapters for the updated prompt as condition.}
    \vspace{-1em}
    \label{fig:overview}
\end{figure*}

\sssection{Video Personalization.} Video personalization aims to integrate the appearance or motion of a designated object into a pre-trained video generation model, enabling the model to reproduce these properties when generating with other prompts. Building upon the progress in the image domain~\cite{dreambooth, iclora, diffusion_self_distillation}, several approaches handle the temporal consistency problem by adding LoRAs~\cite{lora} to the temporal layers of T2V models~\cite{customttt, he2024idanimatorzeroshotidentitypreservinghuman} or learning the motion embeddings from reference videos~\cite{newmove, dreamvideo, motiondirector}. Set-and-Sequence~\cite{set_and_sequence} enables the simultaneous learning of both appearance and motion from a single video by designing the spatio-temporal weight space within the LoRA architecture. Grid-LoRA~\cite{gridlora} further enables reusable video personalization by introducing a grid-based LoRA system that spatially organizes input and output. However, we cannot accurately designate the concept to extract and the way the concepts are combined. The number and type of inputs are also confined, limiting the flexibility of concept composition.

\sssection{Visual Concept Composition.} Composing multiple visual elements from images and videos into a coherent output remains a challenging task. There have been early explorations in decomposing image concepts~\cite{breakascene, conceptexpress, inspirationtree}. Break-A-Scene~\cite{breakascene} relies on explicit mask inputs to achieve concept decomposition, which limits its availability to common users and its ability to extract non-object concepts. Other methods~\cite{conceptexpress, inspirationtree} extract multiple concepts from a single image by jointly learning several tokens, each corresponding to a visual concept. However, the content learned by each token is unpredictable. To achieve concept composition in the image domain, existing works either use explicit spatial conditioning~\cite{omg, yang2024loracomposerleveraginglowrankadaptation, conceptweaver}, which falls short in overlapping or non-object concepts, or fuse multiple LoRAs~\cite{mixofshow, modular_customization, ziplora}, which restricts the type and number of concepts to compose or requires joint optimization among all source images. TokenVerse~\cite{tokenverse} learns a modulation term for each text token to achieve prompt-controlled concept composition. Despite enhanced flexibility, it relies on text-conditioned modulation architectures in DiT~\cite{dit} models, limiting its universality to modern T2V models~\cite{wan, meituanlongcatteam2025longcatvideotechnicalreport}. To extend concept composition to handle videos, previous methods~\cite{dreamvideo, wang2025dualrealadaptivejointtraining, videomage} incorporate dedicated designs to decouple appearance and motion, supporting only the composition of subjects from images and motions from videos. \methodshort ~simultaneously enables complex concept decomposition (non-object concepts and multiple concepts from a single input) and flexible concept composition (selective composition via prompts and composing image and video concepts), offering endless possibilities for visual creators.


\section{Methodology}
\label{sec:method}


\subsection{Overview}
\label{sec:method_overview}





Given $M$ concept images or videos $\{\bm{V}^j_c\}_{j=1}^M$ with their corresponding textual prompt tokens $\{\bm{p}^j_c\}_{j=1}^M$, \methodshort ~aims at composing the visual concepts from the inputs according to the designated prompt $\bm{p}_d$ to generate a coherent visual output $\bm{V}$. 
As illustrated in \cref{fig:overview}, it first learns each text token's corresponding visual appearance or motion via a light-weight binder module for each visual input, and then combines tokens from different source images or videos to generate a target video that composes the individual concepts.
Specifically, during concept binding, a binder structure attached to a DiT-based T2V model~\cite{wan} is utilized to encode the correspondence between visual concepts and textual tokens through one-shot training on different inputs $\{\bm{V}^j_c\}_{j=1}^M$ and $\{\bm{p}^j_c\}_{j=1}^M$ respectively. When integrating concepts from various sources, different parts of the designated prompt $\bm{p}_d$ representing visual concepts are passed through their corresponding binders to compose the updated prompt $\bm{p}_u$, which contains visual concept information and is then fed into DiT blocks to serve as the condition for the composed visual output.


\sssection{Preliminary: Text Conditioning in T2V Models.}
Current mainstream T2V models~\cite{easyanimate, cogvideox, opensora2, hunyuanvideo, hacohen2024ltxvideorealtimevideolatent, wan} adopt the DiT~\cite{dit} architecture with tens of blocks to predict the denoising vector. Each DiT block contains attention layers, an MLP, and a modulation mechanism for the timestep condition. To achieve text conditioning, a prevalent method is to insert a cross-attention layer in each DiT block, which takes the latent tokens $\bm{x}_{in}$ as queries and the textual prompt tokens $\bm{p}$ as keys and values:
\begin{equation}
    \bm{x}_{out}=\texttt{cross\_attention}(\bm{x}_{in},\bm{p},\bm{p}),
    \label{eq:cross_attention}
\end{equation}
where $\bm{x}_{out}$ stands for the updated latent tokens. Through the cross-attention process, the textual information is injected into the DiT model and serves as the condition when predicting the denoising vector.

\subsection{Hierarchical Binder Structure}
\label{sec:adapter_structure}

\begin{figure}[t]
    \centering
    \includegraphics[width=\linewidth]{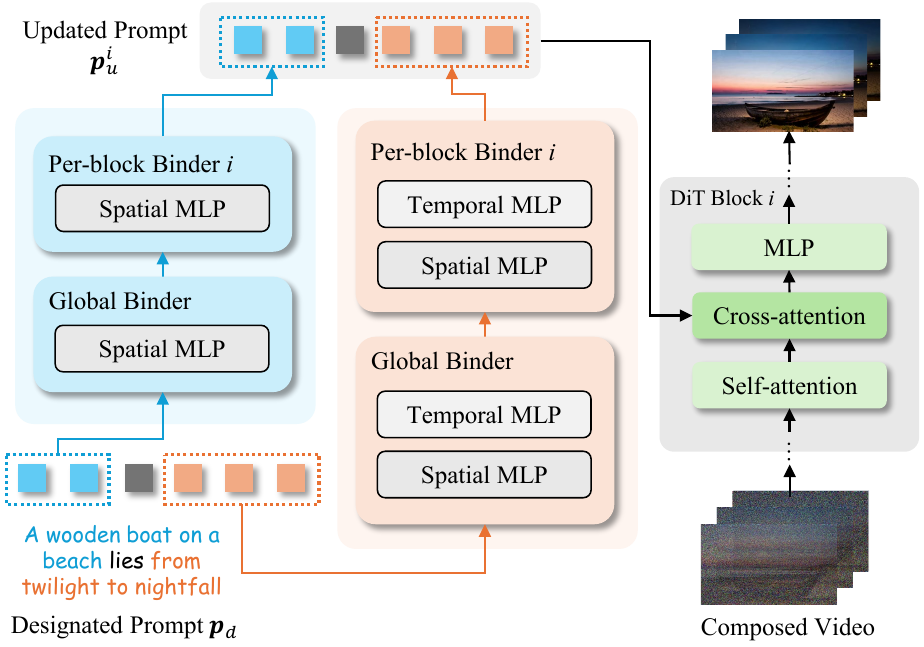}
    \vspace{-1.5em}
    \caption{\textbf{Hierarchical Binder Structure} (\sref{sec:adapter_structure}). It consists of global and per-block binders, where each binder contains an MLP with residual connections. For video inputs, a dual-branch binder structure with spatial and temporal MLPs is incorporated to better address temporal concepts.
    }
    \label{fig:binder}
    \vspace{-1.5em}
\end{figure}


To fully exploit the powerful capability of visual-text association within T2V models for accurate decomposition of complex visual concepts, we attach binders to DiT cross-attention conditioning layers to encode visual concepts into corresponding prompt tokens.
Since DiT blocks have distinct behaviors during the denoising process~\cite{cache_me_if_you_can}, a hierarchical binder structure is designed with a global binder for the overall association and per-block binders for tailored association (\cref{fig:binder}). 
Specifically, each binder $f(\cdot)$ consists of an MLP with a zero-initialized learnable scaling factor $\gamma$ in a residual style, and takes the prompt tokens $\bm{p}$ as input:
\begin{equation}
    f(\bm{p})=\bm{p}+\gamma\cdot\texttt{MLP}(\bm{p}).
    \label{eq:binder}
\end{equation}

For video inputs, a dual-branch binder structure with spatial and temporal MLPs is incorporated to better address temporal concepts (detailed in \sref{sec:tds}). For the training process, the concept prompt tokens $\bm{p}_c$ are first passed through a global binder $f_g(\cdot)$ for a global update to get $\bm{p}_g$. Then, for the \textit{i}-th DiT block, $\bm{p}_g$ are fed into a per-block binder $f_l^i(\cdot)$ to obtain the updated prompt $\bm{p}_u^i$, which are used as the key and value inputs for the cross-attention layer. For the inference process, we first decompose the designated prompt tokens $\bm{p}_d$ according to the correspondence with visual concepts, and then feed each concept-related part into the corresponding binder. Finally, we compose the updated prompt $\bm{p}_u^i$ with the result of each concept binder. This design enables flexible manipulation of visual concepts by composing the designated prompt $\bm{p}_d$.

\sssection{Two-stage Inverted Training Strategy.} Recent studies point out that the denoising process of diffusion models is divided into several stages with different functions~\cite{autodiffusion, perception_prioritized_training}. 
It has been discovered that prioritizing the training on higher noise levels yields better performance~\cite{perception_prioritized_training}. 
To this end, we utilize a two-stage inverted training strategy to enhance the optimization process for hierarchical binders. 
Specifically, we define a noise level threshold $\alpha$ to separate high and low noise levels. In the first stage, we only train the global binder with the probability of $\alpha$ for high noise levels ($\geq\alpha$) and the probability of $1-\alpha$ for low noise levels ($<\alpha$). This setting emphasizes the high noise levels and reduces the optimization steps on low noise levels. In the second training stage, both global and per-block binders are trained without inverting the probability of noise levels.

\subsection{Diversify-and-Absorb Mechanism (DAM)}
\label{sec:dam}

\begin{figure}[t]
    \centering
    \includegraphics[width=\linewidth]{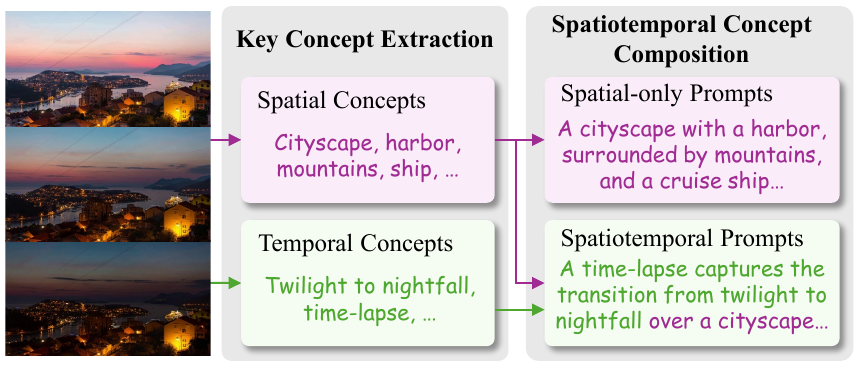}
    \vspace{-1.5em}
    \caption{\textbf{Prompt Diversification} (\sref{sec:dam}). The VLM extracts key spatial and temporal concepts from the visual input, and then composes them into diverse spatial-only or spatiotemporal prompts.
    }
    \label{fig:dam}
    \vspace{-1.5em}
\end{figure}

Establishing accurate concept-token bindings is a notable challenge, especially in one-shot cases. To enable precise association between concepts and prompt tokens in binders, we introduce DAM, which takes advantage of the powerful visual comprehension and reasoning capability of Vision-language Models (VLMs)~\cite{bai2025qwen25vltechnicalreport} to diversify concept prompts while retaining the key conceptual words unchanged during the concept binding process. As shown in \cref{fig:dam}, the prompt diversification process is divided into two stages: key concept extraction and spatiotemporal concept composition. In the key concept extraction stage, the VLM is asked to extract critical concepts from the input image or video and divide them into spatial and temporal concepts. During spatiotemporal concept composition, the VLM composes the extracted concepts into a designated number of diverse prompts with the visual input reference. For images and the first-stage training of videos with a focus on spatial concepts (detailed in \sref{sec:tds}), only spatial concepts are used to form the full prompt. For the second-stage training of videos, the VLM uses both spatial and temporal concepts to generate the complete prompt.

The diversified prompts may not cover all the details in the visual inputs, and those uncovered visual elements can entangle with other prompt tokens, resulting in degraded concept-prompt binding quality. To address this issue, a learnable absorbent token is introduced to minimize the impact of concept-irrelevant details during concept binding by absorbing those distracting visual details. Concretely, when binding the $j$-th visual concept source $\bm{V}_c^j$ with the corresponding textual prompt tokens $\bm{p}_c^j$, we initialize an absorbent token $p_a^j$, and concatenate it with $\bm{p}_c^j$ along the sequence dimension as the input of the hierarchical binder structure. The embeddings of the token $p_a^j$ are updated with other learnable parameters during optimization. When it comes to concept composing, the absorbent token $p_a^j$ is discarded to suppress undesired details.

\subsection{Temporal Disentanglement Strategy (TDS)}
\label{sec:tds}

The ability to compose concepts from both images and videos is of great significance to visual content creators. 
However, significant temporal heterogeneity exists between images and videos~\cite{spatial_temporal_causal_inference}, especially the temporal domain shift caused by the absence of motion in images. This hinders compatibility when directly composing concepts from both sources.
To enable flexible composition of image and video concepts with satisfactory quality, we devise TDS, which aligns the learning paradigm of images and videos by decoupling the training process for video concepts into two stages. In the first stage, we train the binders on individual video frames without temporal concepts in the input prompt. This setting remains the same as the training setting of image concepts, with a focus on binding spatial concepts. In the second stage that trains the binders on full videos and complete prompts with temporal concepts, we adopt a dual-branch binder structure to decouple the learning of spatial and temporal concepts and inherit the knowledge from the first stage. Specifically, we extend the MLP in the original binder with an extra temporal MLP branch $\texttt{MLP}_t$, and then fuse them with a learnable gating module $g(\cdot)$:
\begin{equation}
    \texttt{MLP}(\bm{p})\leftarrow (1-g(\bm{p}))\cdot\texttt{MLP}_s(\bm{p})+g(\bm{p})\cdot\texttt{MLP}_t(\bm{p}),
    \label{eq:spatiotemporal_binder}
\end{equation}
where the weights of $\texttt{MLP}_s$ are taken from the first stage and $g(\cdot)$ is zero-initialized to provide the optimization process with a good initial state. Such a disentanglement strategy alleviates the temporal heterogeneity between images and videos and achieves better results when composing concepts from both images and videos.

\section{Experiments}
\label{sec:exp}

\subsection{Implementation Details}
\label{sec:impl}

We select Wan2.1-T2V-1.3B~\cite{wan} as the base model to apply \methodshort . The MLP structure in binders consists of two linear layers with layer normalization~\cite{ba2016layernormalization} and GELU~\cite{gelu} activation. The binders are trained with a learning rate of $1.0\times 10^{-4}$ for $2400$ iterations per stage. The noise level threshold $\alpha$ in \sref{sec:adapter_structure} is set to $0.875$. We set the length for composed videos to $81$ frames during inference. All other hyperparameters remain the same as Wan2.1~\citep{wan}. Experiments are conducted on NVIDIA RTX 4090 GPUs.

\subsection{Comparisons to Prior Works}
\label{sec:comparison}

To demonstrate the superiority of \methodshort ~over existing visual concept composition works, we conduct quantitative and qualitative comparisons with 4 representative methods: Textual Inversion (Text-Inv)~\cite{textualinversion}, DreamBooth-LoRA (DB-LoRA)~\cite{dreambooth}, DreamVideo~\cite{dreamvideo}, and DualReal~\cite{wang2025dualrealadaptivejointtraining}. We adapt Text-Inv and DB-LoRA on the same T2V model~\cite{wan} as \methodshort ~to support video concepts. Since existing methods that support both images and videos only take one image (subject) and one video (motion) as input, we limit our comparisons to composing concepts from one image and one video for fair comparisons in this section.

\subsubsection{Quantitative Comparisons}
\label{sec:comparison_quantitative}

We construct 40 test cases with images and videos from the DAVIS~\cite{davis} dataset and the Internet for evaluation. Both automatic metrics and human evaluations are adopted for assessing the concept composition performance. For automatic metrics, we use \textit{CLIP-T} to measure the alignment between the generated video and the textual prompt with CLIP~\cite{clip} feature similarities, and choose \textit{DINO-I} to quantify the preservation of visual concepts with the harmonic mean of DINO~\cite{dino} feature similarities between the composed video and all visual inputs. For human evaluations, we asked 28 volunteers to rate the composed video in the following aspects with a 5-point Likert scale: 1) \textit{Concept Preservation}: how well the composed video preserves the concepts from corresponding visual sources. 2) \textit{Prompt Fidelity}: how well the composed video follows the input prompt. 3) \textit{Motion Quality}: the motion quality of the composed video considering motion smoothness, consistency, naturalness, etc. We compute the average score of the 3 aspects as the \textit{Overall Quality}. Please refer to the supplementary for more details on user study settings.

\begin{table}
  \caption{\textbf{Quantitative Comparisons with Prior Arts} (\sref{sec:comparison_quantitative}). Results in \textbf{bold} are the best. \dag~Implemented on Wan2.1~\cite{wan}.}
  \vspace{-0.5em}
  \label{tab:quantitative}
  \centering
  \setlength{\tabcolsep}{1mm}
  \resizebox{\linewidth}{!}{
  \begin{tabular}{l|cc|cccc}
    \toprule
    Method & CLIP-T$\uparrow$ & DINO-I$\uparrow$ & Concept$\uparrow$ & Prompt$\uparrow$ & Motion$\uparrow$ & Overall$\uparrow$ \\
    \midrule
    Text-Inv\textsuperscript{\dag} & 25.96 & 20.47 & 2.14 & 2.17 & 2.94 & \cellcolor{Gray}2.42 \\
    DB-LoRA\textsuperscript{\dag} & 30.25 & 27.74 & 2.76 & 2.76 & 2.51 & \cellcolor{Gray}2.68 \\
    DreamVideo & 27.43 & 24.15 & 1.90 & 1.82 & 1.66 & \cellcolor{Gray}1.79 \\
    DualReal & 31.60 & 32.78 & 3.10 & 3.11 & 2.78 & \cellcolor{Gray}3.00 \\
    \methodshort ~(Ours) & \textbf{32.66} & \textbf{38.04} & \textbf{4.71} & \textbf{4.76} & \textbf{4.46} & \cellcolor{Gray}\textbf{4.64} \\
    \bottomrule
  \end{tabular}
  }
  \vspace{-1.5em}
\end{table}

As displayed in \cref{tab:quantitative}, \methodshort ~consistently outperforms all other methods in both automatic metrics and human evaluations. Compared to the prior art DualReal~\cite{wang2025dualrealadaptivejointtraining}, our method achieves a \textbf{+54.67\%} improvement on the subjective \textit{Overall Quality}. In addition, \methodshort ~also supports the extraction of non-object concepts, learning multiple concepts from a single input, arbitrary image/video input types, and flexible concept composition via prompt manipulation, where previous methods fall short.

\subsubsection{Qualitative Comparisons}
\label{sec:comparison_qualitative}

\begin{figure*}[t]
  \centering
  \includegraphics[width=\linewidth]{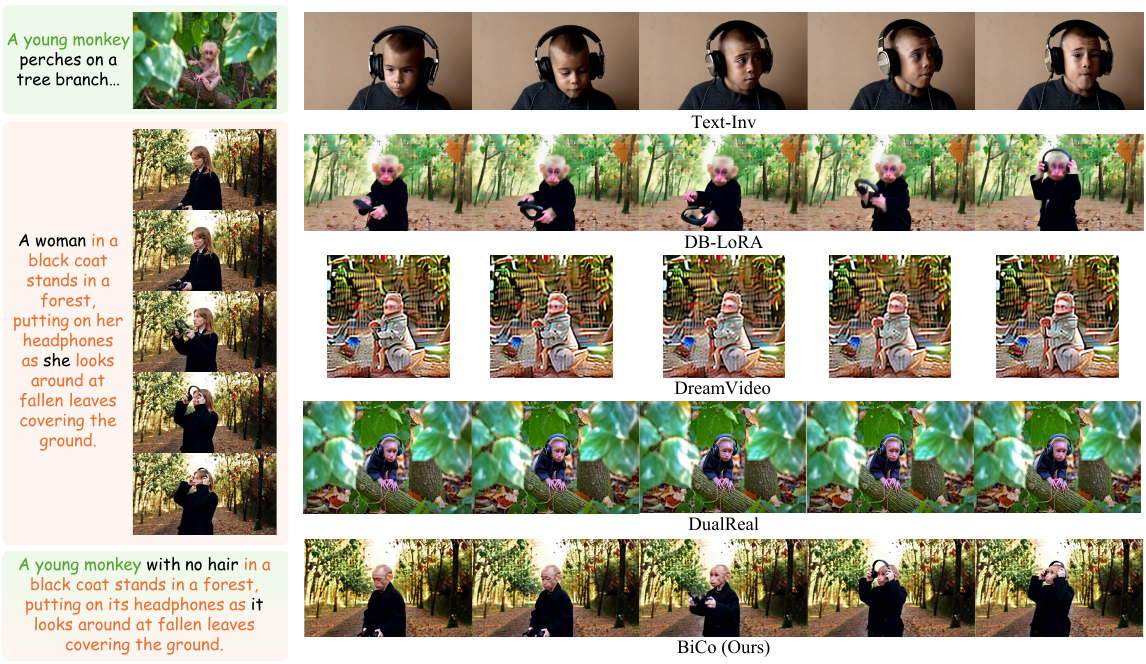}
  \vspace{-1.5em}
  \caption{\textbf{Qualitative Comparisons with Previous Methods} (\sref{sec:comparison_qualitative}). The input visual concepts and composed prompts are on the left.}
  \label{fig:comparison}
  \vspace{-1em}
\end{figure*}

We visualize the composed videos in \cref{fig:comparison} to provide an intuitive comparison with other methods. It shows a creative motion transfer task, where Textual Inversion~\cite{textualinversion} and DreamVideo~\cite{dreamvideo} fails to combine the visual concepts. DualReal~\cite{wang2025dualrealadaptivejointtraining} does not accurately follow the prompt to compose the concepts, and the generated video is almost static. Although DB-LoRA~\cite{dreambooth} mostly follows the designated prompt to integrate visual concepts, there are significant drifts of visual concepts from the original inputs. \methodshort ~best composes the visual concepts according to the given prompt while maintaining the visual concept consistency with the input image and video.


\subsection{Diagnostic Experiments}
\label{sec:diagnostic}

To provide a better understanding of \methodshort 's components, we conduct both quantitative ablations and a case study.


\begin{table}
  \caption{\textbf{Ablations of \methodshort} (\sref{sec:diagnostic}). Results in \textbf{bold} are the best. \TriangleUp ~stands for without two-stage inverted training strategy.}
  \label{tab:ablation}
  \vspace{-0.5em}
  \centering
  \setlength{\tabcolsep}{1mm}
  \resizebox{\linewidth}{!}{
  \begin{tabular}{cccc|cccc}
    \toprule
    Hrc. & Div. & Abs. & TDS & Concept$\uparrow$ & Prompt$\uparrow$ & Motion$\uparrow$ & Overall$\uparrow$ \\
    \midrule
    & & & & 2.16 & 2.60 & 2.26 & \cellcolor{Gray}2.34 \\
    \Checkmark & & & & 2.63 & 2.88 & 2.93 & \cellcolor{Gray}2.81 \\
    \Checkmark & \Checkmark & & & 3.40 & 3.34 & 3.04 & \cellcolor{Gray}3.26 \\
    \Checkmark & \Checkmark & \Checkmark & & 3.55 & 3.43 & 3.43 & \cellcolor{Gray}3.47 \\
    \Checkmark & \Checkmark & & \Checkmark & 3.80 & 3.97 & 3.70 & \cellcolor{Gray}3.82 \\
    \TriangleUp & \Checkmark & \Checkmark & \Checkmark & 2.60 & 2.70 & 2.43 & \cellcolor{Gray}2.58 \\
    \Checkmark & \Checkmark & \Checkmark & \Checkmark & \textbf{4.43} & \textbf{4.47} & \textbf{4.32} & \cellcolor{Gray}\textbf{4.40} \\
    \bottomrule
  \end{tabular}
  }
  \vspace{-1.5em}
\end{table}

\sssection{Quantitative Ablations.} We adopt the human evaluation method in \sref{sec:comparison_quantitative} with another 24 volunteers and the same test cases. The results are presented in \cref{tab:ablation}. 

\begin{figure*}[t]
  \centering
  \includegraphics[width=\linewidth]{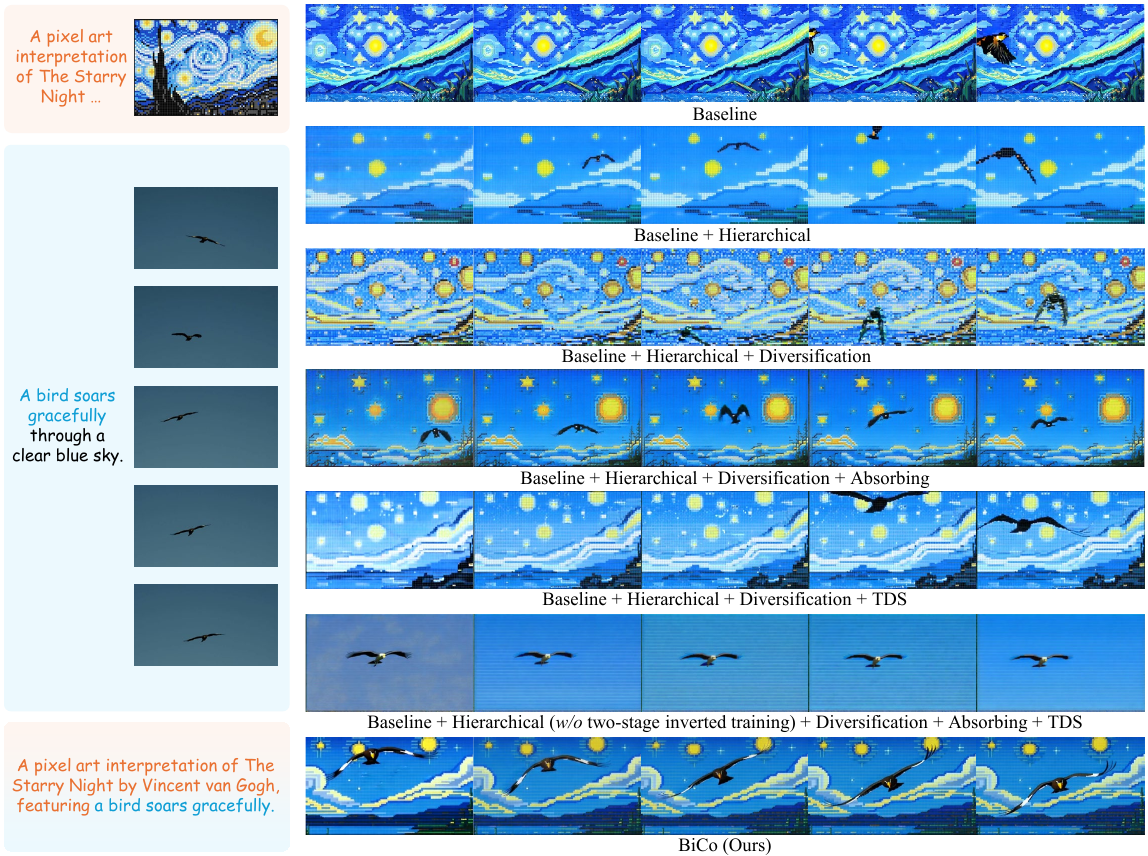}
  \vspace{-1.5em}
  \caption{\textbf{Case Study for Components} (\sref{sec:diagnostic}). The input visual concepts and composed prompts are on the left.
  %
  }
  \vspace{-1em}
  \label{fig:ablation}
\end{figure*}

\sssection{Case Study.} We further illustrate the functions of \methodshort 's components with a concrete visual concept composition sample with an image and a video input in \cref{fig:ablation}. 

\sssection{Baseline.} We start from a simple baseline with only the global binder, omitting the hierarchical design, DAM, and TDS (\#1). This naive baseline method does not achieve satisfactory performance due to limited concept binding capability and image-video compatibility.

\sssection{Hierarchical Binder Structure.} By integrating the hierarchical design of binders (Hrc., \#2), the binding capability of our method is significantly enhanced with per-block binders for tailored concept-token association. The effectiveness is demonstrated by the improvement of \textit{Concept Preservation} and \textit{Motion Quality} in \cref{tab:ablation} and the better reproduction of the bird concept in \cref{fig:ablation} compared to \#1. 

\sssection{Prompt Diversification.} The prompt diversification operation (Div., \#3) enhances the binding accuracy between concepts and prompt tokens under the one-shot training setting of \methodshort . As \cref{tab:ablation} shows, the \textit{Concept Preservation} score rises significantly compared to \#1 with the integration of the prompt diversification operation. However, some unwanted details appear in \cref{fig:ablation}, degrading the composition quality.

\sssection{Absorbent Token.} The absorbent token (Abs.) in AAM facilitates more accurate concept-prompt binding by suppressing prompt-irrelevant details during training, resulting in reduced unwanted elements comparing \#4 to \#3 and \#7 to \#5 in \cref{fig:ablation}. The improvement of \textit{Concept Preservation} and \textit{Motion Quality} in \cref{tab:ablation} further verifies the effectiveness of the absorbent token.

\sssection{TDS.} By decoupling the training process of spatial and temporal concepts in videos, TDS prominently enhances the \textit{Overall Quality} in \cref{tab:ablation} comparing \#5 to \#3 and \#7 to \#4. The qualitative results in \cref{fig:ablation} also improves with better concept detail preservation from both the input image and video. These results validate its effectiveness for improving the compatibility between image and video concepts. 

\sssection{Two-stage Inverted Training Strategy.} The two-stage inverted training strategy plays an essential part in training the hierarchical binders. By first training the global binder with a focus on high noise levels, the strategy provides a better initialization for the full training stage and stabilizes the training process. Without such a training strategy, the optimization becomes hard and unstable, resulting in considerably degraded results in \#6 of both \cref{tab:ablation} and \cref{fig:ablation}.



\subsection{Qualitative Results}
\label{sec:qualitative}

\begin{figure*}[t]
  \centering
  \includegraphics[width=\linewidth]{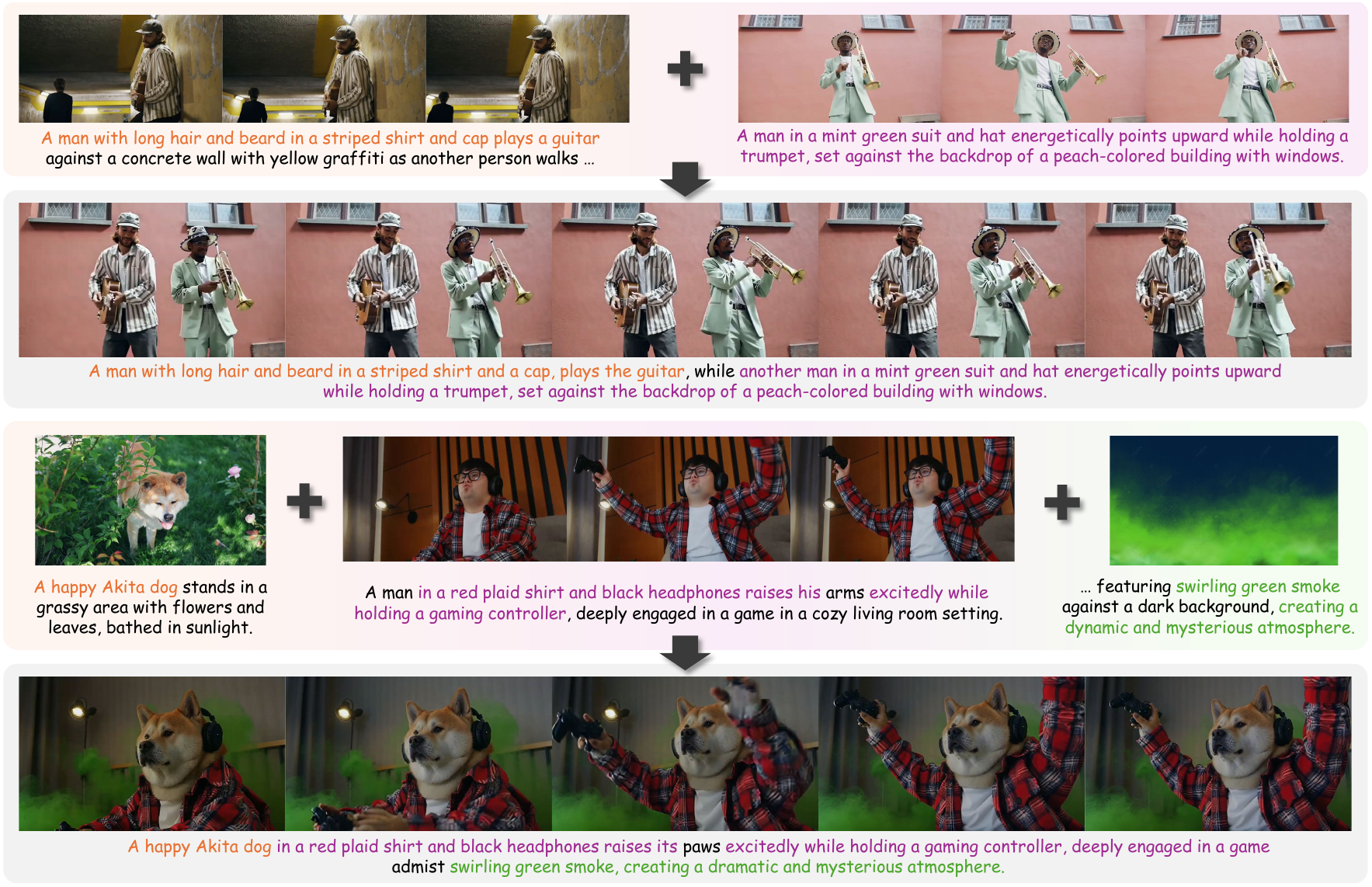}
  \vspace{-1.5em}
  \caption{\textbf{Qualitative Results} (\sref{sec:qualitative}). In each case, the upper row shows the visual inputs, and the lower row presents the composed video.}
  \label{fig:qualitative}
  \vspace{-1em}
\end{figure*}

We present various creative visual concept composition results with \methodshort ~in \cref{fig:teaser,fig:qualitative}, including the composition of non-object concepts (\eg style and motion), and composing multiple visual concepts. As observed, \methodshort ~consistently achieves satisfactory concept consistency, prompt fidelity, and motion quality, validating the superiority of our design. More results can be found in the supplementary.

\subsection{Other Applications}
\label{sec:applications}

\begin{figure}[t]
    \centering
    \includegraphics[width=\linewidth]{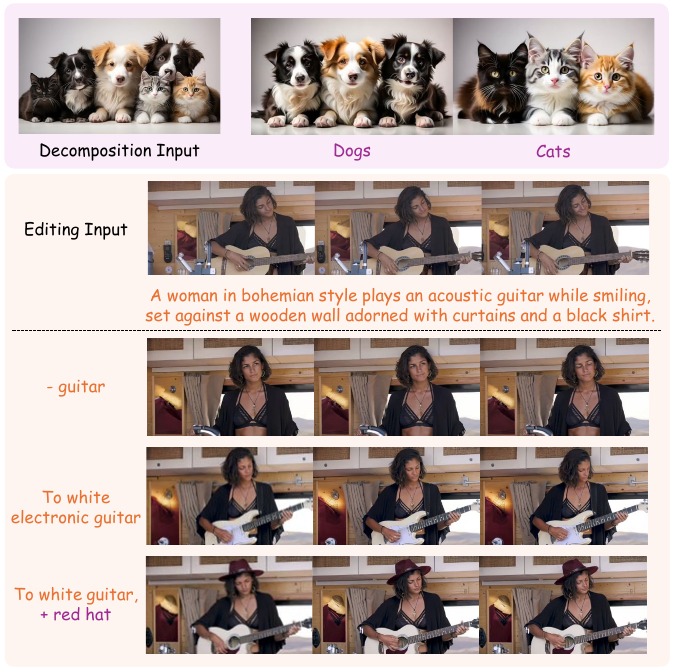}
    \vspace{-1.5em}
    \caption{\textbf{Other Applications} (\sref{sec:applications}). \methodshort ~can also perform other tasks like image/video decomposition and text-guided editing.
    }
    \label{fig:applications}
    \vspace{-1.7em}
\end{figure}

Thanks to the powerful concept binding capability and flexible token manipulation pattern of \methodshort , we can utilize \methodshort ~to perform other creative applications for visual content creation. As the upper part of \cref{fig:applications} illustrates, \methodshort ~possesses the capability of decoupling complex concepts from the visual inputs, such as all the dogs from the input with multiple dogs and cats. This is achieved by keeping only the dog-related tokens in the designated prompt and discarding the cat-related ones when generating the target video with the trained binder. In addition, \methodshort ~can also perform text-guided visual editing, as displayed in the lower part of \cref{fig:applications}. To edit the input image or video, we first perform concept binding and then compose the designated prompt tokens. For the unchanged visual elements, we pass the corresponding prompt tokens through the binder. For the edited parts, the prompt tokens are directly used to compose the designated prompt without updates.

\section{Conclusion and Discussion}
\label{sec:conclusion}

In this work, we propose \methodshort , a one-shot method that can accurately extract complex visual concepts and flexibly combine concepts from both images and videos. It first binds visual concepts with the corresponding prompt tokens and then composes the target prompt with bound tokens from various sources to generate the composed video. It includes a hierarchical binder structure to achieve complex visual concept decomposition, DAM for more accurate concept-token binding, and TDS for enhanced image-video compatibility. Extensive results across various scenarios have validated the effectiveness of \methodshort. We believe that \methodshort ~will boost the community's creativity by providing a handy tool to achieve versatile visual concept composition.

\sssection{Limitations.} \methodshort ~treats each token equally in the concept composition process. Nevertheless, the significance of each token for T2V generation is unevenly distributed. Some tokens that represent subjects and motions play a more important role than the function words. We plan to integrate adaptive designs to highlight those critical tokens in future work. More discussions are included in the supplementary.


\section*{Acknowledgment}

This work was partially supported by a grant from the NSFC/RGC Collaborative Research Scheme Project No. CRS\_HKUST605/25.

{
    \small
    \bibliographystyle{ieeenat_fullname}
    \bibliography{main}
}

\clearpage
\maketitlesupplementary
\appendix

This document includes more details, extra experimental results, corresponding analyses, and further discussions of \methodshort . The document is organized as follows:
\begin{itemize}
    \item \sref{sec:prompts} provides detailed VLM prompts for the prompt diversification process in DAM.
    \item \sref{sec:user_study} gives more details on the user studies.
    \item \sref{sec:abl_training} provides more details on the two-stage inverted training strategy and conducts further ablations.
    \item \sref{sec:analysis_abs} explains the justifications for the absorbent token and provides empirical evidence.
    \item \sref{sec:additional_qualitative_comp} illustrates more qualitative comparisons.
    \item \sref{sec:extra_case} performs another case study to facilitate the understanding of different components of \methodshort .
    \item \sref{sec:discussions} further discusses the limitations with failure cases and the societal impacts of \methodshort .
\end{itemize}
Please refer to the webpage for video results.

\section{Detailed Prompts for DAM}
\label{sec:prompts}

In the prompt diversification process, we utilize a powerful VLM Qwen2.5-VL~\cite{bai2025qwen25vltechnicalreport} to generate diversified concept prompts while retaining the key conceptual words unchanged. During the key concept extraction stage, the VLM is asked to extract essential spatial and temporal concepts from the visual inputs. For image inputs, we use the following textual prompt to extract spatial concepts:

\begin{tcolorbox}[breakable]
You are an image captioning specialist whose goal is to extract the concepts in words or phrases that compose the input image. You need to adhere to the formatting of the examples provided strictly.

\noindent\textbf{Task Requirements}:
\begin{enumerate}
    \item Concepts stand for names of objects, colors, styles, etc;
    \item The overall output should be in English;
    \item The concepts should be brief but concrete, each concept is either a single word or a small phrase. Avoid vague concepts such as "background";
    \item You should be precise and concise;
    \item You should output all the extracted concepts within a "spatial" category as the example.
\end{enumerate}

\noindent\textbf{Example of the concept output}:

\noindent\{``spatial": [``brown cat", ``sunglasses", ``sketch", ``sunny", ``grassland"]\}

\noindent Please output in JSON format (pure text, without markdown formatting).
\end{tcolorbox}

For video inputs, the following textual prompt is adopted to extract both spatial and temporal concepts:

\begin{tcolorbox}[breakable]
You are a video captioning specialist whose goal is to extract the spatial and temporal concepts in words or phrases that compose the input video. You need to adhere to the formatting of the examples provided strictly.

\noindent\textbf{Task Requirements}:
\begin{enumerate}
    \item Spatial concepts stand for names of objects, colors, styles, etc;
    \item Temporal concepts refer to the motion, transitions, and probably viewpoint changes in the video;
    \item The overall output should be in English;
    \item The concepts should be brief but concrete, each concept is either a single word or a small phrase. Avoid vague concepts such as "background";
    \item You should be precise and concise;
    \item You should output all the extracted concepts within a "spatial" category as the example.
\end{enumerate}

\noindent\textbf{Example of the concept output}:

\noindent\{``spatial": [``brown cat", ``sunglasses", ``sketch", ``sunny", ``grassland"], ``temporal": [``jumping", ``running", ``falling", ``gently flowing", ``bright to dark", ``near to far"]\}

\noindent Please output in JSON format (pure text, without markdown formatting).
\end{tcolorbox}

During the spatiotemporal concept composition stage, the VLM is asked to combine the extracted concepts into a number of full prompts according to the visual input. For images and the first-stage training of videos with a focus on spatial concepts, we use the following prompts:

\begin{tcolorbox}[breakable]
You are an image captioning specialist whose goal is to write high-quality English prompts by referring to the extracted concepts and the input image, making them complete and expressive.

\noindent\textbf{Task Requirements}:
\begin{enumerate}
    \item Use the given concepts to describe the image in a concise sentence;
    \item You should make sure that the generated caption matches the image content;
    \item You can rearrange or paraphrase these concepts to form diverse captions;
    \item No matter what language the user inputs, you must always output in English.
\end{enumerate}

\noindent\textbf{Example of the English captions}:
\begin{enumerate}
    \item A boat in a river, with trees and houses on the riverbank, and a foggy sky.
    \item A large brown bear in front of a rocky enclosure. The backdrop features a rustic stone wall and scattered boulders.
    \item A human pose standing with arms crossed in front of a black background.
\end{enumerate}

\noindent Directly output the English caption text.
\end{tcolorbox}

For the second-stage training of videos, the following prompt is adopted:

\begin{tcolorbox}[breakable]
You are a video captioning specialist whose goal is to write high-quality English prompts by referring to the extracted spatial and temporal concepts and the input video, making them complete and expressive.

\noindent\textbf{Task Requirements}:
\begin{enumerate}
    \item Use the given concepts to describe the video in a concise sentence;
    \item You should make sure that the generated caption matches the video content;
    \item You can rearrange or paraphrase these concepts to form diverse captions;
    \item No matter what language the user inputs, you must always output in English.
\end{enumerate}

\noindent\textbf{Example of the English captions}:
\begin{enumerate}
    \item A boat sailing in a river, creating white ripples in the water, with trees and houses on the riverbank, and a foggy sky.
    \item A large brown bear ambles slowly across a rocky enclosure. The backdrop features a rustic stone wall and scattered boulders.
    \item A human pose standing with arms crossed in front of a black background, turning slowly from left to right.
\end{enumerate}

\noindent Directly output the English caption text.
\end{tcolorbox}

\section{User Study Details}
\label{sec:user_study}

We recruited volunteers from various backgrounds to conduct the user study. Each user is given a subset of 10 groups of test cases and is asked to rate the concept consistency, prompt fidelity, and motion quality on a 5-point Likert scale. The detailed questions are as follow:

\begin{itemize}
    \item \textbf{Concept Preservation}: How well do you think that the composed video preserves the concepts from the corresponding visual sources?
    \item \textbf{Prompt Fidelity}: How well do you think that the composed video follows the input prompt?
    \item \textbf{Motion Quality}: Please rate the motion quality of the generated video. You can consider the motion smoothness, consistency, naturalness, etc. Please note that \textbf{still frames without motion are considered low quality}.
\end{itemize}

\section{Extra Details and Ablations on Two-stage Inverted Training Strategy}
\label{sec:abl_training}

The probability distribution for the discretized timestep $t_i$ in inverted training is:
\begin{equation}
    p(t_i)=
    \begin{cases}
    (1-\beta)\cdot\frac{1}{N_{<\alpha}}&,d(t_i)<\alpha \\
    \beta\cdot\frac{1}{N_{\geq\alpha}}&,d(t_i)\geq\alpha
    \end{cases}
    ,
\end{equation}
where $d(t_i)\in[0,1]$ indicates the position of $t_i$ in the scheduler, and $N_*$ is the total number of discretized timesteps in the interval $*$. $\alpha =0.875$ is selected according to the training recipe of Wan2.2 to distinguish higher and lower noise levels. While $\beta$ can be selected in a reasonable range to emphasize the higher noise levels, we empirically found that setting $\beta=\alpha$ exchanges the total probability mass between the higher and lower noise levels and yields satisfactory performance given that higher noise levels originally account for a smaller probability than lower noise levels.

\begin{table}
  \caption{\textbf{Extra Ablations on Two-stage Inverted Training Strategy} (\sref{sec:abl_training}). Results in \textbf{bold} are the best.}
  \label{tab:abl_training}
  \vspace{-0.5em}
  \centering
  \setlength{\tabcolsep}{1mm}
  \resizebox{\linewidth}{!}{
  \begin{tabular}{cc|cccc}
    \toprule
    Two-stage & Inverted & Concept$\uparrow$ & Prompt$\uparrow$ & Motion$\uparrow$ & Overall$\uparrow$ \\
    \midrule
     & & 2.60 & 2.70 & 2.43 & \cellcolor{Gray}2.58 \\
    \Checkmark & & 3.53 & 3.77 & 3.53 & \cellcolor{Gray}3.61 \\
    \Checkmark & \Checkmark & \textbf{4.43} & \textbf{4.47} & \textbf{4.32} & \cellcolor{Gray}\textbf{4.40} \\
    \bottomrule
  \end{tabular}
  }
  \vspace{-1.5em}
\end{table}

We provide additional quantitative ablation results under the same settings in \S\textcolor{cvprblue}{4.3} to facilitate understanding of the two-stage inverted training strategy. Results are shown in \cref{tab:abl_training}, where \textit{Two-stage} means that training the global binder before training the whole hierarchical binder structure, and \textit{Inverted} stands for focusing more on high noise levels in the first stage. We can observe that both techniques are crucial for achieving satisfactory optimization of the binders.

\section{Analysis on Absorbent Token}
\label{sec:analysis_abs}

In T2V models, text tokens are already associated with corresponding visual concepts as a good initial value for further personalization.
This association is the foundation for our binders to learn sample-specific features. With a new absorbent token, it is expected that the model encodes irrelevant information into this token instead of other tokens with good initialization for corresponding visual concepts. The absorbent token is expected not to capture specific concepts, but to prevent other conceptual tokens from being distracted from established initial associations.

\begin{figure}[ht]
  \centering
  \vspace{-0.5em}
  \includegraphics[width=0.95\linewidth]{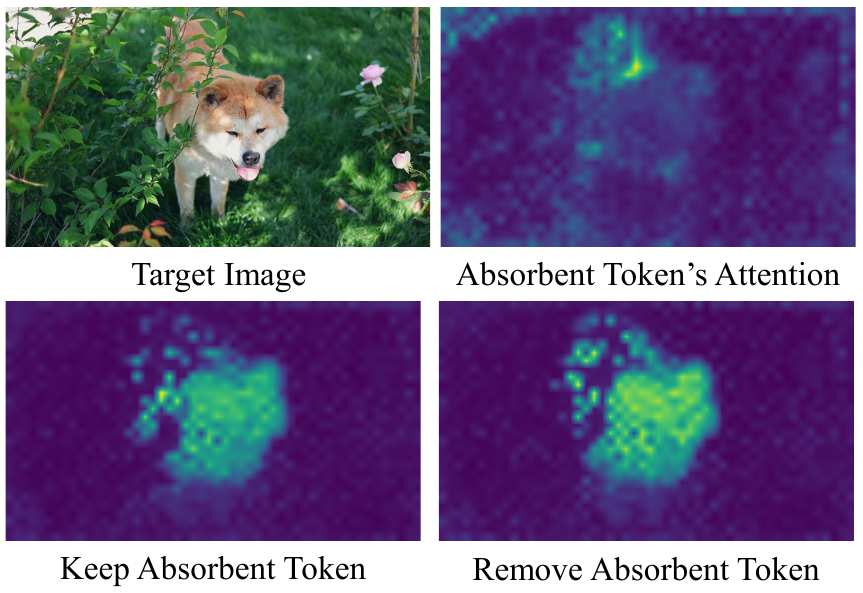}
  \caption{Visualizations of cross-attention maps of target subject tokens (Akita dog) (\sref{sec:analysis_abs}).}
  \label{fig:vis_abs}
  \vspace{-1em}
\end{figure}

We demonstrate the effectiveness of the absorbent token by reconstructing a target image with the trained binder and visualizing the cross-attention maps of the target subject tokens (Akita dog) and the absorbent token in \cref{fig:vis_abs}. As observed, the absorbent token does capture irrelevant details like plants. Removing the trained absorbent token during inference also enhances attention on the target.

\section{Additional Qualitative Comparisons}
\label{sec:additional_qualitative_comp}

\begin{figure*}
  \centering
  \captionsetup[subfigure]{labelformat=empty}
  \begin{subfigure}{\linewidth}
    \centering
    \includegraphics[width=\linewidth]{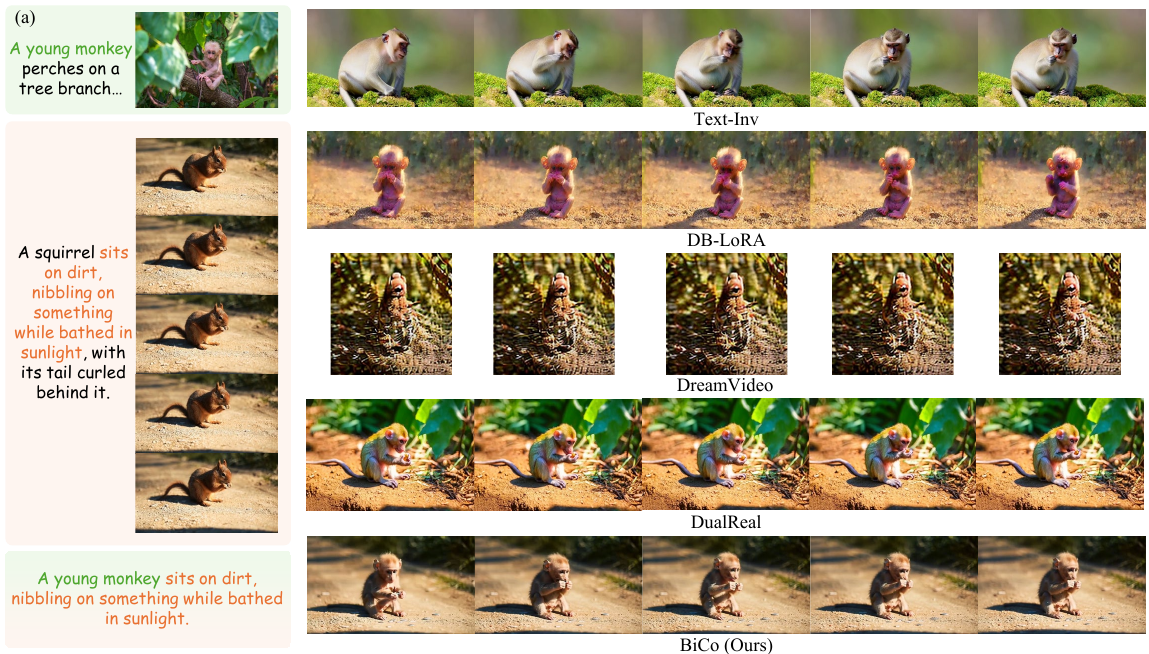}
    \caption{}
    \label{fig:comparison_a}
  \end{subfigure}
  \vfill
  \vspace{-1em}
  \begin{subfigure}{\linewidth}
    \centering
    \includegraphics[width=\linewidth]{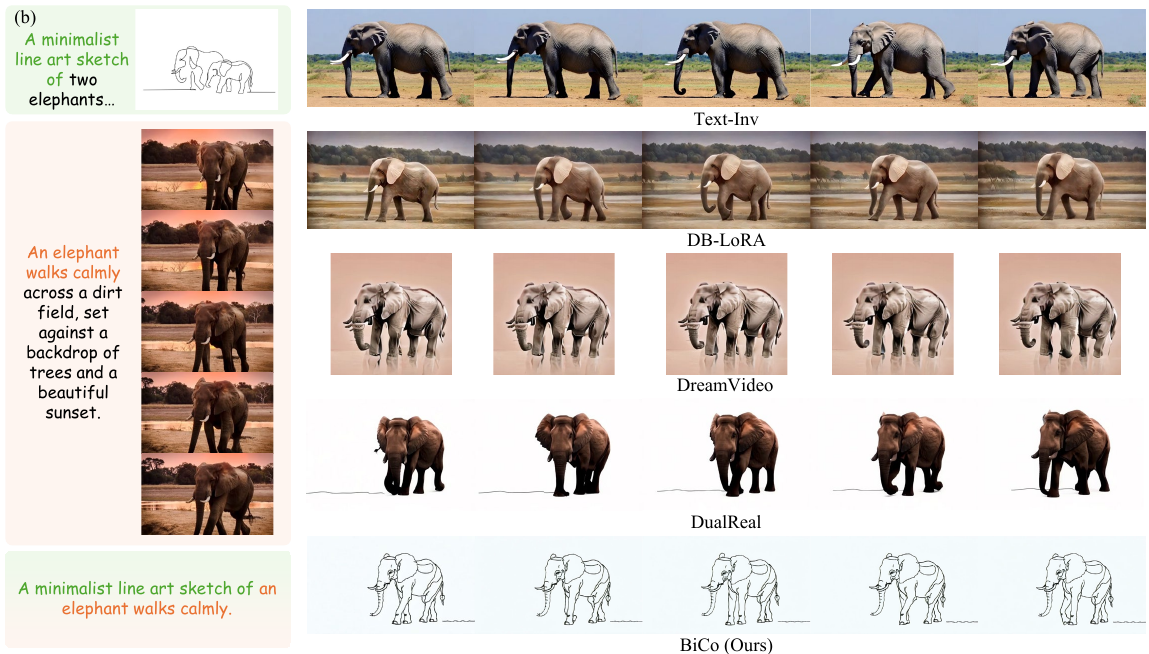}
    \caption{}
    \label{fig:comparison_b}
  \end{subfigure}
  \caption{\textbf{Additional Qualitative Comparisons} (\sref{sec:additional_qualitative_comp}). The input visual concepts and composed prompts are on the left.}
  \label{fig:additional_comparison}
\end{figure*}

We provide more composed videos in \cref{fig:additional_comparison} for additional qualitative comparisons with other methods. \cref{fig:comparison_a} demonstrates a motion transfer task, where Textual Inversion~\cite{textualinversion} and DreamVideo~\cite{dreamvideo} fails to combine the visual concepts. DualReal~\cite{wang2025dualrealadaptivejointtraining} suffers from inadequate visual concept preservation and unintended concept leakage (\eg, the green leaves). Although DB-LoRA~\cite{dreambooth} mostly follows the designated prompt to integrate visual concepts, there are significant drifts of visual concepts from the original inputs (\eg, the direction of the squirrel). \methodshort ~achieves the best result in composing the visual concepts according to the given prompt while maintaining the consistency of visual concepts with the input image and video.

\cref{fig:comparison_b} illustrates a creative style transfer task to integrate the line art sketch style with the subject in a video. All previous methods~\cite{textualinversion, dreambooth, dreamvideo, wang2025dualrealadaptivejointtraining} fail in this task to learn and compose the style concept. This sample further verifies the flexible versatile controllability of \methodshort .

\section{Extra Case Study}
\label{sec:extra_case}

\begin{figure*}
  \centering
  \includegraphics[width=\linewidth]{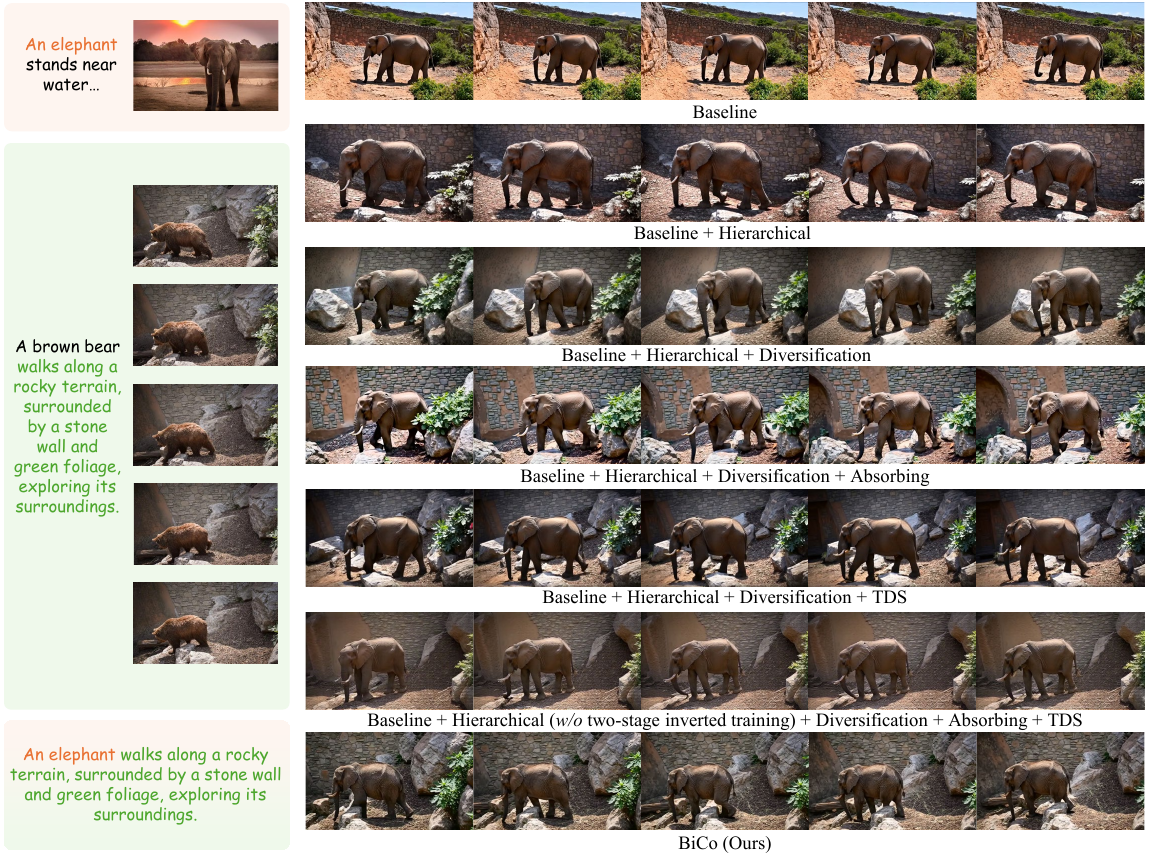}
  \caption{\textbf{Extra Case Study for Components} (\sref{sec:extra_case}). The input visual concepts and composed prompts are on the left.
  }
  \label{fig:ablation_supp}
\end{figure*}

We further illustrate the functions of \methodshort 's components with another concrete visual concept composition sample in \cref{fig:ablation_supp}. Comparing \#2 to \#1, we can observe that the hierarchical binder structure enables our method to encode more visual information into binders, resulting in better concept preservation results. The prompt diversification operation (\#3) and the absorbent token (\#4) in DAM enhance the accuracy of concept-prompt binding, better preserving background details in the composed videos. The effectiveness of the absorbent token can also be verified by the enhanced background preservation in \#7 compared to \#5. TDS further improves the composition quality by enhancing the compatibility between image and video concepts, as illustrated by comparing \#7 to \#4 and \#5 to \#3. The two-stage inverted training strategy significantly stabilizes the optimization process, bringing considerably better results in the same optimization steps (\#7 to \#6). The video results can be found in the webpage.



\section{More Discussions}
\label{sec:discussions}

\subsection{Limitations}
\label{sec:limitations}

\begin{figure*}
  \centering
  \includegraphics[width=\linewidth]{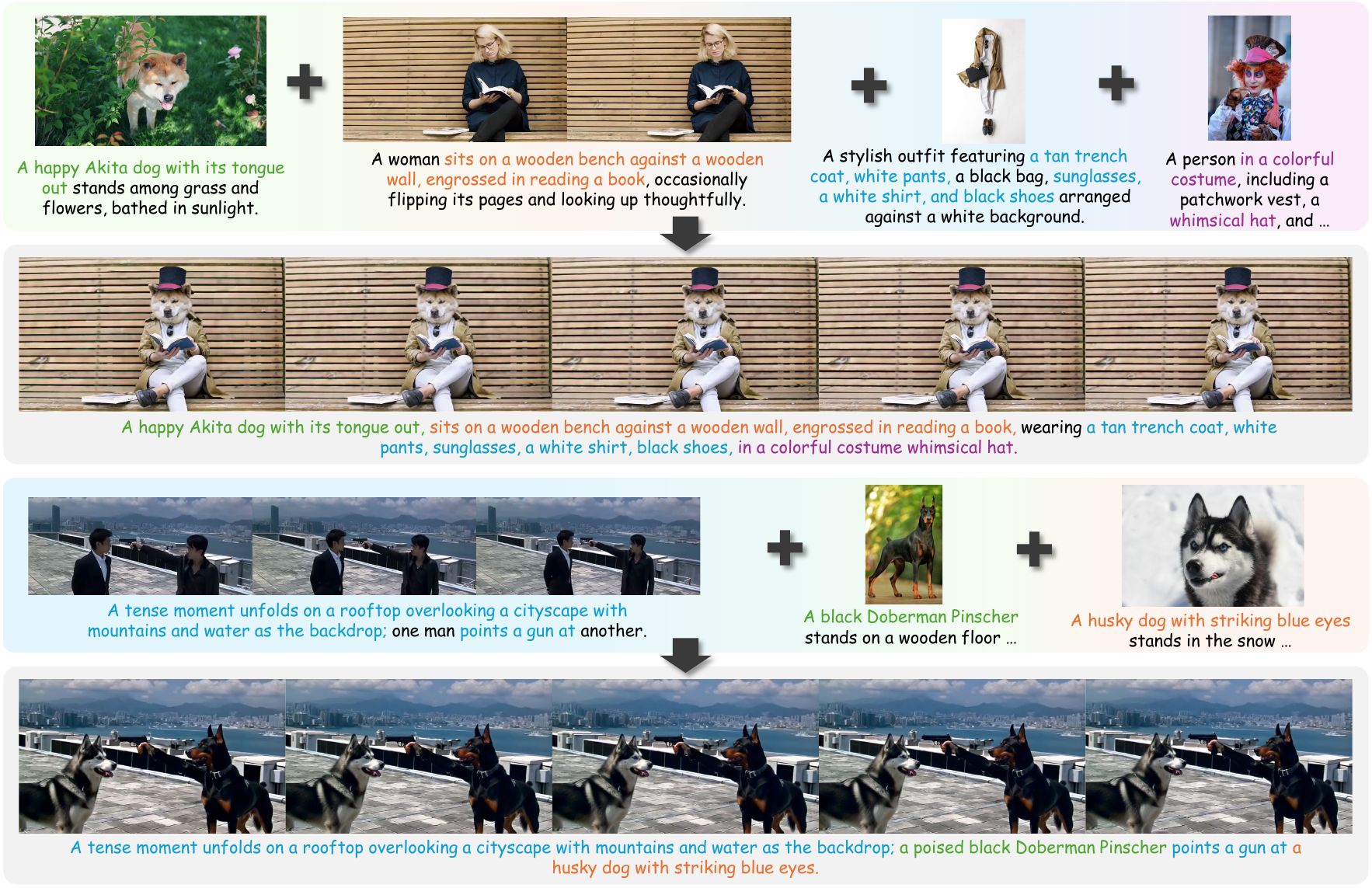}
  \caption{\textbf{Failure Cases} (\sref{sec:limitations}). In each case, the upper row shows the visual inputs, and the lower row presents the composed video.}
  \label{fig:failure}
\end{figure*}

The significance of each prompt token for T2V generation is unevenly distributed. Some tokens that represent subjects and motions play a more important role than the function words. In addition, when a concept is visually complex or deviates significantly from the \textit{average looking} of the text token, the binder's representation capability for each token may be insufficient to accommodate all the visual information. Nevertheless, \methodshort ~treats each token equally in the concept composition process, which can result in unintended concept drifts. For instance, in the upper part of \cref{fig:failure}, \methodshort ~fails to accurately reproduce the colorful whimsical hat in the composed video, where the hat's appearance differs considerably from an average hat. We plan to integrate adaptive designs to highlight critical tokens in our future work.

Furthermore, \methodshort ~also falls short when the composition requires some common sense reasoning. For example, the composed video in the lower part of \cref{fig:failure} simply adds an additional leg to the Doberman Pinscher to hold the gun instead of raising an existing leg, resulting in a total of 5 legs in a single dog. This issue may be alleviated by integrating the strong reasoning capabilities of VLMs to design a more comprehensive captioning and composing paradigm.


\subsection{Societal Impacts}
\label{sec:impacts}

\methodshort ~enables flexible visual concept composition for both images and videos through a one-shot paradigm, enabling practitioners to experiment with visual concepts from multiple sources to implement their creativity. For individual creators, the one-shot nature of our method allows them to integrate AI-assisted visual content composition into their workflows without extensive training. For commercial teams, our method provides them with a new opportunity to flexibly combine their intermediate results and other assets, boosting the novelty of the produced visual content.

On the other hand, with \methodshort's powerful capability to manipulate visual concepts, it can be used to produce fabricated images and videos that appear highly realistic, posing significant challenges for verifying the authenticity of visual media. Such content can distort public perception and raise privacy concerns when fake contents featuring an individual are generated in an unauthorized way.


\end{document}